\pdfoutput=1

\documentclass[11pt]{article}

\usepackage[]{EMNLP2022}

\usepackage{times}
\usepackage{latexsym}

\usepackage[T1]{fontenc}
\usepackage[utf8]{inputenc}
\usepackage{microtype}
\usepackage{inconsolata}
\usepackage{amssymb}
\usepackage{amsmath}
\usepackage{multirow}
\usepackage{graphicx}
\usepackage{makecell}
\usepackage{todonotes} 
\usepackage{CJK}

\title{VarMAE: Pre-training of Variational Masked Autoencoder for Domain-adaptive Language Understanding}

\author{Dou Hu$^{1,2,3}$\thanks{{\ \ }This work was done when the author was at Ping An.} ,
        Xiaolong Hou$^{3}$,
        Xiyang Du$^{3}$,
        Mengyuan Zhou$^{3}$, \\
        {\bf Lianxin Jiang$^{3}$, 
        Yang Mo$^{3}$, 
        Xiaofeng Shi$^{3}$ } \\
        $^1$ Institute of Information Engineering, Chinese Academy of Sciences  \\
        $^2$ School of Cyber Security, University of Chinese Academy of Sciences \\
        $^3$ Ping An Life Insurance Company of China, Ltd. \\
        \texttt{hudou@iie.ac.cn, \{houxiaolong430, duxiyang037, zhoumengyuan425,} \\ \texttt{jianglianxin769, moyang853, shixiaofeng309\}@pingan.com.cn} \\
}

\begin{document}
\maketitle
\begin{abstract}
Pre-trained language models have achieved promising performance on general benchmarks, but underperform when migrated to a specific domain. Recent works perform pre-training from scratch or continual pre-training on domain corpora. However, in many specific domains, the limited corpus can hardly support obtaining precise representations. To address this issue, we propose a novel Transformer-based language model named VarMAE for domain-adaptive language understanding. Under the masked autoencoding objective, we design a context uncertainty learning module to encode the token's context into a smooth latent distribution. The module can produce diverse and well-formed contextual representations. Experiments on science- and finance-domain NLU tasks demonstrate that VarMAE can be efficiently adapted to new domains with limited resources. 

\end{abstract}
\section{Introduction}
Pre-trained language models (PLMs) have achieved promising performance in natural language understanding (NLU) tasks on standard benchmark datasets \cite{DBLP:conf/emnlp/WangSMHLB18,DBLP:conf/coling/XuHZLCLXSYYTDLS20}.
Most works \cite{DBLP:conf/naacl/DevlinCLT19,DBLP:journals/corr/abs-1907-11692} leverage the Transformer-based pre-train/fine-tune paradigm to learn contextual embedding from large unsupervised corpora. 
Masked autoencoding, also named masked language model in BERT \cite{DBLP:conf/naacl/DevlinCLT19}, is a widely used pre-training objective that randomly masks tokens in a sequence to recover. 
The objective can lead to a deep bidirectional representation of all tokens in a BERT-like architecture. 
However, these models, which are pre-trained on standard corpora (e.g., Wikipedia), tend to underperform when migrated to a specific domain due to the \textit{distribution shift} \cite{DBLP:journals/bioinformatics/LeeYKKKSK20}. 

Recent works perform pre-training from scratch \cite{, DBLP:journals/health/GuTCLULNGP22,yao2022nlp} or continual pre-training \cite{DBLP:conf/acl/GururanganMSLBD20,DBLP:conf/iclr/WuCLLQH22} on large domain-specific corpora.
But in many specific domains (e.g., finance), effective and intact unsupervised data is difficult and costly to collect due to data accessibility, privacy, security, etc.
The limited domain corpus may not support pre-training from scratch \cite{DBLP:conf/emnlp/ZhangRSCFFKRSW20}, and also greatly limit the effect of continual pre-training due to the \textit{distribution shift}.
Besides, some scenarios (i.e., non-industry academics or professionals) have limited access to computing power for training on a massive corpus.
Therefore, how to obtain effective contextualized representations from the limited domain corpus remains a crucial challenge.

Relying on the distributional similarity hypothesis \cite{DBLP:journals/corr/abs-1301-3781}
in linguistics, 
that similar words have similar contexts, 
masked autoencoders (MAEs) leverage co-occurrence between the context of words to learn word representations.
However, when pre-training on the limited corpus, most word representations can only be learned from fewer co-occurrence contexts, leading to sparse word embedding in the semantic space.
Besides, in the reconstruction of masked tokens, it is difficult to perform an accurate point estimation \cite{DBLP:conf/emnlp/LiGLPLZG20}  based on the partially visible context for each word.
That is, the possible context of each token should be diverse.
The limited data only provides restricted context information, which causes MAEs to learn a relatively poor context representation in a specific domain.

To address the above issue, we propose a novel \textbf{Var}iational \textbf{M}asked \textbf{A}uto\textbf{e}ncoder (\textbf{VarMAE}), a regularized version of MAEs, for a better domain-adaptive language understanding. 
Based on the vanilla MAE, we design a context uncertainty learning (CUL) module for learning a precise context representation when pre-training on a
limited corpus.
Specifically, the CUL encodes the token's point-estimate context in the semantic space into a smooth latent distribution.
And then, the module reconstructs the context using feature regularization specified by prior distributions of latent variables.
In this way, latent representations of similar contexts can be close to each other and vice versa \cite{DBLP:conf/emnlp/LiHNBY19}.
Accordingly, we can obtain a smoother space and more structured latent patterns.

We conduct continual pre-training on unsupervised corpora in two domains (science and finance) and then fine-tune on the corresponding downstream NLU tasks.
The results consistently show that VarMAE outperforms representative language models including vanilla pre-trained \cite{DBLP:journals/corr/abs-1907-11692} and continual pre-training methods \cite{DBLP:conf/acl/GururanganMSLBD20}, when adapting to new domains with limited resources.
Moreover, compared with masked autoencoding in MAEs, the objective of VarMAE can produce a more diverse and well-formed context representation.

\section{VarMAE}
\begin{figure}[t]
    \centering
    \includegraphics[width=0.98\linewidth]{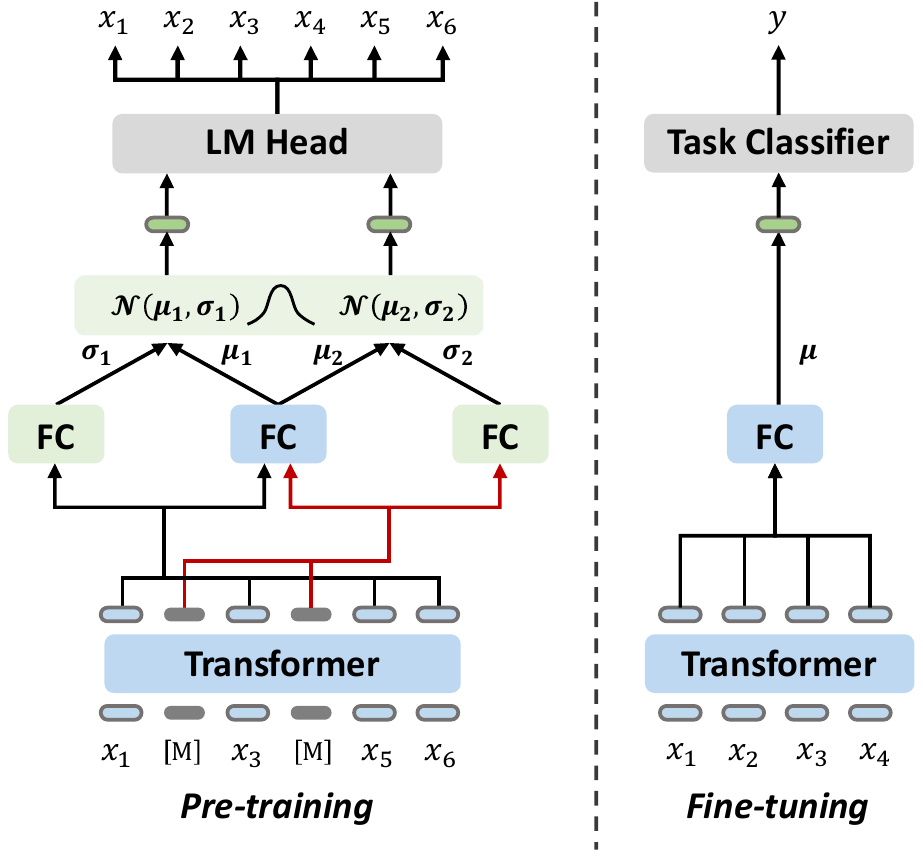} 
    \caption{The architecture of VarMAE. Based on the vanilla MAE, a CUL module is used to learn diverse and well-formed context representations for all tokens. }
    \label{fig:model}
\end{figure}

In this section, we develop a novel Variational Masked Autoencoder (VarMAE) to improve vanilla MAE for domain-adaptive language understanding. The overall architecture is shown in Figure~\ref{fig:model}. Based on the vanilla MAE, we design a context uncertainty learning (CUL) module for learning a precise context representation when pre-training on a limited corpus.

\subsection{Architecture of Vanilla MAE}

\paragraph{Masking}

We randomly mask some percentage of the input tokens and then predict those masked tokens.
Given one input tokens $X=\{x_1,...,x_n\}$ and $n$ is the sentence length, the model selects a random set of positions (integers between 1 and $n$) to mask out $M=\{m_1,...,m_k\}$, where $k=\lceil0.15n\rceil$ indicates 15\% of tokens are masked out.
The tokens in the selected positions are replaced with a $\mathtt{[MASK]}$ token. 
The masked sequence can be denoted as $X^{\text{masked}} = \text{REPLACE}(X, M, \mathtt{[MASK]})$.

\paragraph{Transformer Encoder}
{Vanilla MAE usually} adopts a multi-layer bidirectional Transformer \cite{DBLP:conf/nips/VaswaniSPUJGKP17} as basic encoder like previous pre-training model  \cite{DBLP:journals/corr/abs-1907-11692}.
Transformer can capture the contextual information for each token in the sentence via self-attention mechanism, and generate a sequence of contextual embeddings. Given the masked sentence $X^{\text{masked}}$, the context representation is denoted as $\mathbf{C}=\{\mathbf{c}_1,...,\mathbf{c}_N\}$.

\paragraph{Language Model Head}
We adopt the language model (LM) head to predict the original token based on the reconstructed representation. The number of output channels of LM head equals the number of input tokens. 
Based on the context representation $\mathbf{c}_i$, the distribution of the masked prediction is estimated by:
   $
   p_{\theta}(\mathbf{x}_i | \mathbf{c}_i) = softmax(\mathbf{W} \mathbf{c}_i + \mathbf{b}),
   $
where $\mathbf{W}$ and $\mathbf{b}$ denote the weight matrices of one fully-connected layer. $\theta$ refers to the trainable parameters.
The predicted token can be obtained by $x' = arg\max_i p_{\theta}(\mathbf{x}_i | \mathbf{c}_i)$, where $x'$ denotes the predicted original token.

\subsection{Context Uncertainty Learning} \label{sec:cul}
Due to the flexibility of natural language, one word may have different meanings under different domains.
In many specific domains, the limited corpus can hardly support obtaining precise representations.
To address this, we introduce a context uncertainty learning (CUL) module to learn regularized context representations for all tokens.
These tokens include masked tokens with more noise and unmasked tokens with less noise.
Inspired by variational autoencoders (VAEs) \cite{DBLP:journals/corr/KingmaW13,DBLP:conf/iclr/HigginsMPBGBML17}, 
we use latent variable modeling techniques to quantify the \textit{aleatoric uncertainty}\footnote{The aleatoric uncertainty, or data uncertainty, is the uncertainty that captures noise inherent in the observations.}
\cite{der2009aleatory,abdar2021review} of these tokens.

Let us consider the input token $x$ is generated with an unobserved continuous random variable $\mathbf{z}$.
We assume that $x_i$ is generated from a conditional distribution $p_{\theta}(\mathbf{x} | \mathbf{z})$, where $\mathbf{z}$ is generated from an isotropic Gaussian prior distribution 
$p_{\theta}(\mathbf{z}) =
\mathcal{N}(\mathbf{z}; \boldsymbol{0}, \mathbf{I})$.
To learn the joint distribution of the observed variable $x$ and its latent variable factors $\mathbf{z}$, the optimal objective is to maximize the marginal log-likelihood of $x$ in expectation over the whole distribution of latent factors $\mathbf{z}$:
\begin{equation}
    \max_{\theta} \mathbb{E}_{p_{\theta}(\mathbf{z}) }
    [  \log p_\theta (\mathbf{x} | \mathbf{z})  ]. \label{eq:1}
\end{equation}

Since masked and unmasked tokens have relatively different noise levels, the functions to quantify the \textit{aleatoric uncertainty} of these two types should be different.
{We take CUL for masked tokens as an example.}
Given each input masked token $x_i^m$ and its corresponding context representation $\mathbf{c}^m_i$,
the true posterior $p_{\theta}(\mathbf{z}^m | {x}^m_i)$ is approximated as  $p_{\theta'}(\mathbf{z}^m | \mathbf{c}^m_i)$ due to the distributional similarity hypothesis \cite{DBLP:journals/corr/abs-1301-3781}.
Inspired by \citet{DBLP:journals/corr/KingmaW13},
we assume $p_{\theta'}(\mathbf{z}^m | \mathbf{c}^m_i)$ takes on an approximate Gaussian form with a diagonal covariance, 
and let the variational approximate posterior
be a multivariate Gaussian with a diagonal covariance structure.
This variational approximate posterior is denoted as $q_{\phi}(\mathbf{z}^m | \mathbf{c}^m_i)$:
\begin{equation}
q_{\phi}(\mathbf{z}^m | \mathbf{c}^m_i) = \mathcal{N}(\mathbf{z}^m; \boldsymbol{\mu}^m_i, {\boldsymbol{\sigma}^m_i}^2 \mathbf{I}), \label{eq:2}
\end{equation}
where $\mathbf{I}$ is diagonal covariance, $\phi$ is the variational parameters.
Both parameters (mean as well as variance) are input-dependent and  predicted by MLP (a fully-connected neural network with a single hidden layer), i.e., $\boldsymbol{\mu}^m_i = f_{\phi_{\mu}}(\mathbf{c}^m_i)$, $\boldsymbol{\sigma}^m_i = f_{\phi_{\sigma}}(\mathbf{c}^m_i)$, where $\phi_\mu$ and $\phi_\sigma$ refer to the model parameters respectively w.r.t output $\boldsymbol{\mu}^m_i$ and ${\boldsymbol{\sigma}}^m_i$.
Next, we sample a variable $\mathbf{z}_i^m$ from the approximate posterior $q_{\phi}(\mathbf{z}^m | \mathbf{c}^m_i)$, and then feed it into the LM head to predict the original token.

Similarly, CUL for each unmasked token $x^{\bar{m}}_i$ adopts in a similar way and samples a latent variable $z^{\bar{m}}_i$ from the variational approximate posterior $q_{\phi}(\mathbf{z}^{\bar{m}} | \mathbf{c}^{\bar{m}}_i) = \mathcal{N}(\mathbf{z}^{\bar{m}}; \boldsymbol{\mu}^{\bar{m}}_i, {\boldsymbol{\sigma}^{\bar{m}}_i}^2 \mathbf{I})$,  where ${\mu}^{\bar{m}}_i$ and  $\boldsymbol{\sigma}^{\bar{m}}_i$ are predicted by MLP.

In the implementation, we adopt $f_{\phi_{\mu}}$ with shared parameters to obtain $\boldsymbol{\mu}^m$ and $\boldsymbol{\mu}^{\bar{m}}$.
Conversely, two $f_{\phi_{\sigma}}$ with independent parameters are used to obtain 
$\boldsymbol{\sigma}^m$ and $\boldsymbol{\sigma}^{\bar{m}}$, for $x^{m}$ with more noise and $x^{\bar{m}}$ with less noise, respectively.
After that, batch normalization \cite{DBLP:conf/icml/IoffeS15} is applied to avoid the \textit{posterior collapse}\footnote{The posterior collapse, or KL vanishing, is that the decoder in VAE learns to reconstruct data independent of the latent variable $\mathbf{z}$, and the KL vanishes to $0$.} \cite{DBLP:conf/acl/ZhuBLMLW20}.
By applying the CUL module, the context representation is not a deterministic point embedding any more, but a stochastic embedding sampled from $\mathcal{N}(\mathbf{z}; \boldsymbol{\mu}, {\boldsymbol{\sigma}}^2 \mathbf{I})$ in the latent space.
Based on the reconstructed representation, the LM head is adopted to predict the original token.

\subsection{Training Objective}
To learn a smooth space where latent representations of similar contexts are close to each other and vice versa, the objective function is:
\begin{equation}
\begin{split}
    \max_{\phi, \theta} \mathbb{E}_{
    {x} \sim \mathbf{D}
    }[\mathbb{E}_{
    \mathbf{z} \sim q_{\phi}(\mathbf{z} | \mathbf{c})
    } 
    [\log p_\theta (\mathbf{x} | \mathbf{z})]],
    \\ 
    \text{s.t.} \ D_{KL}( 
            q_{\phi}(\mathbf{z} | \mathbf{c}) 
            \|
            p_{\theta}(\mathbf{z}))  < \delta,
\end{split}
\end{equation}
where $\delta>0$ is a constraint, and 
$q_\phi(\mathbf{z} | \mathbf{c}) $ is the variational approximate posterior of the true posterior $p_{\theta}(\mathbf{z} | {x})$ (see Section~\ref{sec:cul}).
$D_{KL}(\cdot)$ denotes the KL-divergence term, which serves as the regularization that forces prior distribution $p_{\theta}$ to approach the approximated posterior $q_{\phi}$.
Then, for each input sequence, the loss function is developed as a weighted sum of loss functions for masked tokens $\mathcal{L}^{{m}}$ and unmasked tokens $\mathcal{L}^{\bar{m}}$.
The weights are normalization factors of masked/unmasked tokens in the current sequence. 
\begin{equation}
  \resizebox{0.89\linewidth}{!}{$
\begin{split}      
& \ \mathcal{L}^{\tau} =  \mathbb{E}_{ \mathbf{z}^{\tau} \sim q_{\phi}(\mathbf{z}^{\tau} | \mathbf{c}^{\tau}) }[\log p_\theta (\mathbf{x}^{\tau} | \mathbf{z}^{\tau})]        \\
&  - \lambda^{\tau}  D_{KL}( q_{\phi}(\mathbf{z}^{{\tau}} | \mathbf{c}^{{\tau}}) \| p_{\theta}(\mathbf{z}^{{\tau}}) ), \tau \in \{ m, \bar{m}\},
  \\ 
\end{split}
$}
\end{equation}
where $\lambda^{m}$ and $\lambda^{\bar{m}}$ are trade-off hyper-parameters.
Please see Appendix~\ref{app:loss} for more details.

As the sampling of $\mathbf{z}_i$ is a stochastic process, we use  \textit{re-parameterization} trick \cite{DBLP:journals/corr/KingmaW13} to make it trainable:
    $\mathbf{z}_i =  \boldsymbol{\mu}_i +  \boldsymbol{\sigma}_i  \odot \epsilon,  \epsilon \sim \mathcal{N}(0, \mathbf{I}),
    $
where $\odot$ refers to an element-wise product.
Then, KL term  $D_{KL}(\cdot)$  is computed as:
\begin{equation}
\resizebox{0.89\linewidth}{!}{$
\begin{split}
    D_{KL}( 
            q_{\phi} &
            (\mathbf{z}  | \mathbf{c}) 
            \| 
           p_{\theta}(\mathbf{z})
           )  =   -\frac{1}{2} (1 + \log \boldsymbol{\sigma}^2 - \boldsymbol{\mu}^2 - \boldsymbol{\sigma}^2  ). 
\end{split}
$}
\end{equation}

For all tokens, the CUL forces the model to be able to reconstruct the context using feature regularization specified by prior distributions of latent variables.
Under the objective of VarMAE, latent vectors with similar contexts are encouraged to be smoothly organized together.
After the pre-training, we leverage the Transformer encoder and $f_{\phi_{\mu}}$ to fine-tune on downstream tasks.

\begin{table*}[t]
    \centering
      \resizebox{0.9\linewidth}{!}{$
    \begin{tabular}{l|c|c|c|c|c|c|c|c|c|c} 
    \hline
    \multicolumn{1}{c|}{\multirow{3}{*}{Model}} 
    & \multicolumn{5}{c|}{{\textit{Science-domain}}}
    & \multicolumn{5}{c}{{\textit{Finance-domain}}}    \\ \cline{2-11}
    & \textit{ACL-ARC}  & \textit{SciCite}    & \textit{JNLPBA}   & \textit{EBM-NLP}  &  \multirow{2}{*}{{Avg.}}
    & \textit{OIR}      & \textit{MTC}      & \textit{IEE}           & \textit{PSM}      & \multirow{2}{*}{{Avg.}} 
    \\ \cline{2-5} \cline{7-10}
    &  \multicolumn{2}{c|}{CLS}     
    &   NER      
    &   SE
    & 
    &  \multicolumn{2}{c|}{CLS}              
    &   NER      
    &   TM              
    &   \\    \hline 
RoBERTa	
&      74.58	&		84.85	&		73.09	&		75.11	&		76.91	
&	   66.64	&		54.95	&		67.77	&		46.65	&		59.00	\\	
TAPT
&      68.10	&		86.23	&		72.54		&		74.09		&		75.24		
&		65.16	&		53.18	&	68.80		&	49.71    	&		59.21    \\ 
DAPT
&      70.02	&		84.20	&		73.85	&		75.88	&		75.99	
&	   65.54	&		54.49	&		65.90	&		46.47	&		58.10 	\\	
VarMAE
&      \textbf{76.50}	&		\textbf{86.32}	
&	   \textbf{74.43}	&		\textbf{76.01}
&	   \textbf{78.32}	
&	   \textbf{68.77}	&		\textbf{56.58}	
&	   \textbf{70.15}	&		\textbf{53.68}	
&	   \textbf{62.30} 
\\       \hline
    \end{tabular}
    $}
    \caption{Results on 
    science- and finance-domain downstream tasks.
    All compared pre-trained models are fine-tuned on the task dataset.
    For each dataset, we run three random seeds and report the average result of the test sets.
    We report the micro-average F1 score for CLS and TM, entity-level F1 score for NER, and token-level F1 score for SE.
    Best results are highlighted in bold.}
    \label{tab:results}
\end{table*}
\section{Experiments}
We conduct experiments on science- and finance-domain NLU tasks to evaluate our method.

\subsection{Domain Corpus and Downstream Tasks}

\begin{table}[t]
    \centering
          \resizebox{0.95\linewidth}{!}{
    \begin{tabular}{c|c|c|c|c} 
    \hline
      \multicolumn{1}{c|}{\multirow{2}{*}{Corpus Size}}   & \multicolumn{2}{c|}{\multirow{1}{*}{\it Science-domain}}   & \multicolumn{2}{c}{\multirow{1}{*}{\it  Finance-domain}}   \\    \cline{2-5} 
    & \multicolumn{1}{c|}{DAPT}  & \multicolumn{1}{c|}{VarMAE}
    & \multicolumn{1}{c|}{DAPT}  & \multicolumn{1}{c}{VarMAE} \\ \hline
    $|\mathcal{D}|/3$            
        &  76.77       &  77.82                            
        &  59.56    &   62.04                           \\   
    $|\mathcal{D}|$ 
        & 75.99     &  78.32                
        & 58.10     &  62.30                            \\  \hline 
    \end{tabular}
    }
    \caption{Average results on all downstream tasks against different corpus sizes of pre-training.
    $|\mathcal{D}|$ is the corpus size for corresponding domain.}
    \label{tab:pretrain_corpus}
\end{table}

\begin{table}[t]
    \centering
          \resizebox{0.95\linewidth}{!}{
    \begin{tabular}{c|p{2.6cm}<{\centering}|p{2.6cm}<{\centering}}
    \hline
      \multicolumn{1}{c|}{Masking Ratio}   & \multirow{1}{*}{\it Science-domain}   & \multirow{1}{*}{\it Finance-domain}   \\    \hline 
    5\%                                     &  77.27                            &  58.54                               \\   
    \multirow{1}{*}{15\%}                   &  78.32                            &  62.30                            \\  
    30\%                                    &  76.95                            &  59.12                             \\  \hline
    \end{tabular}
    }
    \caption{Average results of VarMAE on all downstream tasks against different masking ratios of pre-training.
    }
    \label{tab:mask}
\end{table}

\begin{table*}[t]
    \centering
          \resizebox{\linewidth}{!}{
    \begin{tabular}{c|p{0.37\linewidth}|c|c|c|c}
    \hline
      \multicolumn{1}{c|}{\multirow{2}{*}{No.}}  & \multicolumn{1}{c|}{\multirow{2}{*}{Example}}  
      & \multicolumn{1}{c|}{\multirow{2}{*}{Gold}}  & \multicolumn{1}{c|}{Pred.}
      & \multicolumn{1}{c|}{Pred.}  & \multicolumn{1}{c}{Pred.}  \\ 
      & &  \multicolumn{1}{c|}{} & \multicolumn{1}{c|}{(RoBERTa)} 
      & \multicolumn{1}{c|}{(DAPT)} & \multicolumn{1}{c}{(VarMAE)} \\ \hline
      \multirow{1}{*}{1} 
      & {\it Can forearm superficial injury insure accidental injury?}  
      & \multicolumn{1}{c|}{\multirow{2}{*}{ 
      \makecell[c]{
      {\it Accident}\begin{CJK}{UTF8}{gbsn} \footnotesize{{(意外)}}\end{CJK}; \\ {{\it Disease underwriting} \begin{CJK}{UTF8}{gbsn}\footnotesize {(疾病核保)} \end{CJK}}}}
      }
      & \multicolumn{1}{c|}{\multirow{1}{*}{\it Disease underwriting}}
      & \multicolumn{1}{c|}{\multirow{1}{*}{\it Accident}}  
      &  \multicolumn{1}{c}{\multirow{2}{*}{\makecell[c]{{\it Accident}; \\ \it Disease underwriting}}}
      \\ 
      & {\begin{CJK}{UTF8}{gbsn}\footnotesize (前臂浅表损伤是否投保意外保险？) \end{CJK}} 
      &  &  &  \\  \hline
        \multirow{1}{*}{2} 
      & \textit{Medical demands inspire quality care.} 
      &  \multicolumn{1}{c|}{\multirow{2}{*}{
      \makecell[c]{\begin{CJK}{UTF8}{gbsn}{{\it Pension} {\footnotesize (养老)};} \end{CJK} \\
      \begin{CJK}{UTF8}{gbsn}{{\it Risk education} {\footnotesize (风险教育)}} \end{CJK}
      } 
      }}
      &  \multicolumn{1}{c|}{\multirow{1}{*}{\it Pension}}
      &  \multicolumn{1}{c|}{\multirow{1}{*}{\it Pension}}
      &  \multicolumn{1}{c}{\multirow{2}{*}{
      \makecell{{\it Pension}; \\ \it Risk education}}
      } \\
      &   {\begin{CJK}{UTF8}{gbsn}\footnotesize (医疗需求激发品质养老。) \end{CJK}} &  &  &  \\   \hline
    \multirow{1}{*}{3} 
      & \textit{How does high incidence cancer protection calculate the risk insurance?} 
      & \multicolumn{1}{c|}{\multirow{2}{*}{
      \makecell[c]{\begin{CJK}{UTF8}{gbsn}{{\it Critical illness} {\footnotesize (重疾)};} \end{CJK} \\
      \begin{CJK}{UTF8}{gbsn}{{\it Insurance rules} {\footnotesize (投保规则)}} \end{CJK} }
      }}
      &  \multicolumn{1}{c|}{\multirow{1}{*}{\it Insurance rules}}
      &  \multicolumn{1}{c|}{\multirow{1}{*}{\it Insurance rules}}
      & \multicolumn{1}{c}{\multirow{2}{*}{
      \makecell{{\it Critical illness};\\ \it Insurance rules}}}
    \\ 
    &  {\begin{CJK}{UTF8}{gbsn}\footnotesize (高发癌症保障计划如何计算风险保额？) \end{CJK}}   &  &  &   \\  \hline
      \multirow{1}{*}{4} 
      & \textit{What are the features of ABC Comprehensive Care Program? }
        & \multicolumn{1}{c|}{\multirow{2}{*}{
        \makecell{\begin{CJK}{UTF8}{gbsn}{{\it Product introduction} {\footnotesize{(产品介绍)}};} \end{CJK} \\
      \begin{CJK}{UTF8}{gbsn}{{\it Critical illness} {\footnotesize{(重疾)}}} \end{CJK} }
        }}
      & \multicolumn{1}{c|}{\multirow{1}{*}{\it Product introduction}}
      & \multicolumn{1}{c|}{\multirow{1}{*}{\it Product introduction}}
      & \multicolumn{1}{c}{\multirow{1}{*}{\it Product introduction}} \\
    &    {\begin{CJK}{UTF8}{gbsn}\footnotesize (ABC全面呵护计划特色包括什么内容？) \end{CJK}} &  &  &  \\
      \hline
    \end{tabular}
    }
    \caption{
    Case studies in the multi-label topic classification (MTC) task of financial business scenarios. The table shows four examples of spoken dialogues in the test set, their gold labels and predictions by three methods (RoBERTa, DAPT and VarMAE). We translate original Chinese to English version for readers.    }
    \label{tab:case}
\end{table*}

\paragraph{Domain Corpus}
For science domain,  we collect 0.6 million English abstracts (0.1B tokens) of computer science and broad biomedical fields, which are sampled from Semantic Scholar corpus \cite{DBLP:conf/naacl/AmmarGBBCDDEFHK18}.
For finance domain, we collect 2 million cleaned Chinese sentences (0.3B tokens) from finance-related online platforms (e.g., \textit{Sina Finance}\footnote{\url{https://finance.sina.com.cn/}}, \textit{Weixin Official Account Platform}\footnote{\url{https://mp.weixin.qq.com/}}, and \textit{Baidu Zhidao}\footnote{\url{https://zhidao.baidu.com/}}) 
and business scenarios\footnote{\url{https://life.pingan.com/}\label{code-life}}.
The 1 million sentences in this corpus are from finance news, sales/claims cases, product introduction/clauses, and finance encyclopedia entries, while the remaining 1 million sentences are collected from the internal corpus and log data in business scenarios.
\paragraph{Downstream Tasks and Datasets}
We experiment with four categories of NLP downstream tasks: text classification (CLS), named entity recognition (NER), span extraction (SE), and text matching (TM).
For science domain, we choose four public benchmark datasets: ACL-ARC \cite{DBLP:journals/tacl/JurgensKHMJ18} and SciCite \cite{DBLP:conf/naacl/CohanAZC19} for citation intent classification task, JNLPBA \cite{DBLP:conf/bionlp/CollierK04} for bio-entity recognition task, EBM-NLP \cite{DBLP:conf/acl/NenkovaLYMWNP18} for PICO extraction task.
For finance domain, we choose four real-world  financial business datasets\textsuperscript{\ref{code-life}}: OIR for outbound intent recognition task, MTC for multi-label topic classification task, IEE for insurance-entity extraction task, and PSM for pairwise search match task. 
The details of datasets are included in Appendix~\ref{sec:appendix_dataset}.

\subsection{Experimental Setup}
We compare VarMAE with the following baselines:
\textbf{RoBERTa} \cite{DBLP:journals/corr/abs-1907-11692}
is an optimized BERT with a masked autoencoding objective, and is to directly fine-tune on given downstream tasks.
\textbf{TAPT} \cite{DBLP:conf/acl/GururanganMSLBD20} is a continual pre-training model on a task-specific corpus.
\textbf{DAPT} \cite{DBLP:conf/acl/GururanganMSLBD20} is a continual pre-training model on a domain-specific corpus.

Experiments are conducted under PyTorch\footnote{\url{https://pytorch.org/}} and using 2/1 NVIDIA Tesla V100 GPUs with 16GB memory for pre-training/fine-tuning. 
During pre-training, we use 
\textit{roberta-base}\footnote{ \url{https://huggingface.co/}\label{code}} and
\textit{chinese-roberta-wwm-ext}\textsuperscript{\ref{code}} to initialize the model for science (English) and finance domains (Chinese), respectively.
During the pre-training of VarMAE, we freeze the embedding layer and all layers of Transformer encoder to avoid catastrophic forgetting \cite{DBLP:conf/nips/French93,DBLP:conf/emnlp/ArumaeSB20} of previously general learned knowledge. And then we optimize other network parameters (e.g., the LM Head and CUL module) by using Adam optimizer \cite{DBLP:journals/corr/KingmaB14} with the learning rate of $5e^{-5}$. 
The number of epochs is set to 3.
We use gradient accumulation step of 50 to achieve the large batch sizes (i.e., the batch size is 3200). The trade-off co-efficient $\lambda$ is set to 10 for both domains selected from $\{1, 10,  100 \}$.
For fine-tuning on downstream tasks, most hyperparameters are the same as in pre-training,  except for the following settings due to the limited computation. The batch size is set to 128 for OIR, and 32 for other tasks.
The maximum sequence length is set to 64 for OIR, and 128 for other tasks.
The number of epochs is set to 10.
More details are listed in Appendix~\ref{sec:appendix_exp}.

\subsection{Results and Analysis}
Table~\ref{tab:results} shows the results on science- and finance-domain downstream tasks.
In terms of the average result, VarMAE yields 1.41\% and 3.09\% absolute performance improvements over the best-compared model on science and finance domains, respectively. 
It shows the superiority of domain-adaptive pre-training with context uncertainty learning.
DAPT and TAPT obtain inferior results. 
It indicates that the small domain corpus limits the continual pre-training due to
the \textit{distribution shift}.

We report the average results on all tasks against different corpus sizes of pre-training in Table~\ref{tab:pretrain_corpus}  {(see Appendix~\ref{sec:appendix_corpus} for details)}. 
VarMAE consistently achieves better performance than DAPT even though a third of the corpus is used.
When using full corpus, DPAT's performance decreases but VarMAE's performance increases, which proves our method has a promising ability to adapt to the target domain with a limited corpus.

Table~\ref{tab:mask} shows the average results of VarMAE on all tasks against different masking ratios of pre-training {(see Appendix~\ref{sec:appendix_mask} for details)}. 
Under the default masking strategies\footnote{{
80\% for replacing the target token with $\mathtt{[MASK]}$ symbol, 10\% for keeping the target token as is, and 10\% for replacing the target token with another random token.}}, 
the best masking rate is 15\%, which is the same as BERT and RoBERTa.

\subsection{Case Study}
As shown in Table~\ref{tab:case}, we randomly choose several samples from the test set in the multi-label topic classification (MTC) task. 

For the first {case}, RoBERTa and DAPT each predict one label correctly. 
It shows that both general and domain language knowledge have a certain effect on the domain-specific task.
However, none of them identify all the tags completely.
This phenomenon reflects that the general or limited continual PLM is not sufficient for the domain-specific task.
For the second and third {cases}, these two comparison methods cannot classify the topic label \textit{Risk education} and \textit{Critical illness}, respectively.
It indicated that they perform an isolated point estimation and have a relatively poor context representation.
Unlike other methods, our VarMAE can encode the token's context into a smooth latent distribution and produce diverse and well-formed contextual representations.
As expected, VarMAE predicts the first three examples correctly with limited resources.

For the last case, all methods fail to predict \textit{Critical illness}. 
We notice that \textit{ABC Comprehensive Care Program} is a product name related to critical illness insurance. Classifying it properly may require some domain-specific structured knowledge. 

\section{Conclusion}

We propose a novel Transformer-based language model named VarMAE for domain-adaptive language understanding with limited resources. A new CUL module is designed to produce a diverse and well-formed context representation. Experiments on science- and finance-domain tasks demonstrate that VarMAE can be efficiently adapted to new domains using a limited corpus. 
Hope that VarMAE can guide future foundational work in this area.

\section*{Limitations}
All experiments are conducted on a small pre-training corpus  due to the limitation of computational resources. The performance of VarMAE pre-training on a larger corpus  needs to be further studied. 
Besides, VarMAE cannot be directly adapted to downstream natural language generation tasks since our model does not contain a decoder for the generation. 
This will be left as future work. 

\vspace{-0.5mm}
\section*{Acknowledgements}
This research is supported by Ping An Life Insurance. We thank the reviewers for their insightful and constructive comments.

\clearpage
\appendix

\section*{Appendix Overview}

In this supplementary material, we provide: 
(i) the related work, 
(ii) objective derivation of the proposed VarMAE,
(iii) detailed description of experimental setups,
(iv) detailed results,  
and (v) our contribution highlights.

\section{Related Work}
\begin{table*}[t]
    \centering
    \resizebox{\linewidth}{!}{
    \begin{tabular}{c|l|l|r|r|r|r|c|r|l}
    \hline
    \multicolumn{2}{c|}{Dataset Name}   & \multicolumn{1}{c|}{Task Name}   
                                  &  \multicolumn{1}{c|}{Train}     &  \multicolumn{1}{c|}{Dev}   & \multicolumn{1}{c|}{Test} & \multicolumn{1}{c|}{\# Entities}
                                  &  \multicolumn{1}{c|}{Avg/Min/Max}       
                                  &  \multicolumn{1}{c|}{Class}  
                                  &  \multicolumn{1}{c}{Source} \\ \hline 
    \multirow{4}{*}{\rotatebox{90}{\it Science}} &
    ACL-ARC 
    & Citation Intent Classification    &  	1,688    & 	114	    &  139	    & -& 42/4/224     & 	6	
    &  NLP field \\ 
   & SciCite 
    &	Citation Intent Classification	        & 	7,320    & 	916     & 	1,861   &  -& 34/7/228     & 	3
    & Multiple scientific fields \\
    & JNLPBA 
    & Bio-entity Recognition            &   16,807&	1,739	& 3,856	& 59,963 &27/2/204      &   5
    &	Biomedical field     \\     
    & EBM-NLP     
    & PICO Extraction                   & 27,879 &	7,049 &	2,064  & 77,360& 37/1/278     &   3   
    & Clinical medicine field \\
    \hline
       \multirow{4}{*}{\rotatebox{90}{\it Finance}} 
    &OIR     & Outbound Intent Recognition & 36,885 & 9,195 & 3,251 & - & 16/2/69 & 34 & F1, F2 \\   
   & MTC     & Multi-label Topic Classification  & 66,670 & 2,994 & 4,606 & - & 15/2/203 & 39 
  & F1, F2, F3, F4  
    \\  
   & IEE      & Insurance-entity Extraction           & 19,136  &   4,784  &   19,206  &  13,128 & 21/1/388  &   2 
  & F1, F2  \\  
& \multirow{1}{*}{PSM}     & \multirow{1}{*}{Pairwise Search Match}
                             & \multirow{1}{*}{11,812}     &  \multirow{1}{*}{1,476}  & \multirow{1}{*}{1,477}
                             &  \multirow{1}{*}{-} & 7/2/100; 14/1/134  
                             & \multirow{1}{*}{4} 
                             & \multirow{1}{*}{F1, F2}       \\ 
                            \hline
    \end{tabular}
    }
    \caption{Dataset statistics of science- and finance-domain downstream tasks. Avg, Min, and Max indicate the average, minimum, and maximum length of sentences, respectively. ``Class"  refers to the number of classes. 
    F1, F2, F3 and F4 mean the insurance, sickness, job and legal fields, respectively.
    }
    \label{tab:datasets_sci_fi}
\end{table*}

\begin{table}[t]
    \centering
    \resizebox{0.98\linewidth}{!}{
    \begin{tabular}{p{0.58\linewidth}|p{0.4\linewidth}<{\centering}}
    \hline
    \multicolumn{1}{c|}{\multirow{1}{*}{\bf Hyperparameter}}
    & \multicolumn{1}{c}{\bf Assignment} \\  \hline
    Number of Epoch                          &      3  \\
    Trade-off Weight $\lambda$                              &      10  \\ 
    Number of Layers                              & 12       \\
    Hidden size                                             & 768        \\
    FFN inner hidden size                              & 3072        \\
    Attention heads                                  &   12       \\
    Attention head size                                &   64      \\
    Dropout                                         &  0.1          \\
    Attention Dropout                             &  0.1        \\
    Peak Learning Rate                         &  $5e^{-5}$   \\ 
    Maximum Length                             &  128       \\
    Batch Size                                            &  64       \\
    Gradient Accumulation Steps  & 50  \\
    Optimization Steps      &  \{504, 1830\}      \\ 
    Weight Decay              &  0.0     \\ 
    Adam $\epsilon$               &  $1e^{-6}$   \\
    Adam $\beta_1$         & 0.9        \\
    Adam $\beta_2$                                   & 0.98      \\ \hline 
    \end{tabular}
    }
    \caption{Hyperparameters for pre-training on a domain-specific corpus for each domain.
    The optimization steps are 504 and 1830 for science- and finance-domain, respectively.}
    \label{tab:app_params}
\end{table}

\begin{table}[t]
    \centering
     \resizebox{0.98\linewidth}{!}{
    \begin{tabular}{p{0.58\linewidth}|p{0.4\linewidth}<{\centering}}
    \hline
    \multicolumn{1}{c|}{\multirow{1}{*}{\textbf{Hyperparameter}}}   &   \textbf{Assignment}  \\ \hline 
    Number of Epoch         &  10   \\ 
    Maximum Length          &  \{64, 128\}      \\
    Batch Size              &   \{32, 128\}        \\   
    Learning Rate           & \multicolumn{1}{c}{$5e^{-5}$}    \\   
    Dropout                 &   0.1               \\ 
    Weight Decay            &   \multicolumn{1}{c}{0.0} \\   
    Warmup ratio            &   
    \multicolumn{1}{c}{0.06} \\  
    \hline 
    \end{tabular}
    }
    \caption{Hyperparameters for fine-tuning on science- and finance-domain downstream tasks. 
    The maximum sequence length is set to 64 for OIR, and is set to 128 for other tasks.
    The batch size is set to 128 for OIR, and is set to 32 for other tasks.
    }
    \label{tab:app_params_fine}
\end{table}

\subsection{General PLMs}
Traditional works \cite{DBLP:conf/nips/MikolovSCCD13,DBLP:conf/emnlp/PenningtonSM14}
represent the word as a single vector representation, which cannot disambiguate the word senses based on the surrounding context.
Recently, unsupervised pre-training on large-scale corpora significantly improves performance, either for Natural Language Understanding (NLU) \cite{DBLP:conf/naacl/PetersNIGCLZ18,DBLP:conf/naacl/DevlinCLT19,DBLP:journals/taslp/CuiCLQY21} or for Natural Language Generation (NLG) \cite{DBLP:journals/jmlr/RaffelSRLNMZLL20,DBLP:conf/nips/BrownMRSKDNSSAA20,DBLP:conf/acl/LewisLGGMLSZ20}. 
Following this trend, considerable progress \cite{DBLP:journals/corr/abs-1907-11692,DBLP:conf/nips/YangDYCSL19,DBLP:conf/iclr/ClarkLLM20,DBLP:journals/tacl/JoshiCLWZL20,DBLP:conf/iclr/0225BYWXBPS20,DBLP:conf/emnlp/DiaoBSZW20} has been made to boost the performance via improving the model architectures or exploring novel pre-training tasks.
Some works \cite{DBLP:journals/corr/abs-1904-09223,DBLP:conf/acl/ZhangHLJSL19,DBLP:conf/acl/QinLT00JHS020} enhance the model by integrating structured knowledge from external knowledge graphs.

Due to the flexibility of natural language, one word may have different meanings under different domains. These methods underperform when migrated to specialized domains. Moreover, simple fine-tuning \cite{DBLP:conf/acl/RuderH18,DBLP:conf/seke/HuW20,DBLP:conf/pkdd/WeiHZTZWHH20,hu2022pali} of PLMs is also not sufficient for domain-specific applications.

\subsection{Domain-adaptive PLMs}
Recent works perform pre-training from scratch \cite{ DBLP:journals/health/GuTCLULNGP22,yao2022nlp} or continual pre-training \cite{alsentzer2019publicly,DBLP:journals/corr/abs-1904-05342,DBLP:journals/bioinformatics/LeeYKKKSK20,DBLP:conf/acl/GururanganMSLBD20,DBLP:conf/iclr/WuCLLQH22,DBLP:conf/acl/QinZLL0SZ22} on domain-specific corpora.

Remarkably, \citet{DBLP:conf/emnlp/BeltagyLC19,DBLP:conf/emnlp/ChalkidisFMAA20} 
explore different strategies to adapt to new domains, including pre-training from scratch and further pre-training.
\citet{DBLP:conf/lrec/BoukkouriFLZ22} find that both of them perform at a similar level when pre-training on a specialized corpus, but the former 
requires more resources. 
\citet{yao2022nlp} jointly optimize the task and language modeling objective from scratch.
\citet{DBLP:conf/emnlp/ZhangRSCFFKRSW20,DBLP:conf/emnlp/Tai0DCK20,DBLP:conf/acl/YaoHWDW21} extend the vocabulary of the LM with domain-specific terms for further gains.
\citet{DBLP:conf/acl/GururanganMSLBD20} show that domain- and task-adaptive pre-training methods can offer gains in specific domains.
\citet{DBLP:conf/acl/QinZLL0SZ22} present an efficient lifelong pre-training method for emerging domain data.

In most specific domains, collecting large-scale corpora is usually inaccessible. 
The limited data makes pre-training from scratch infeasible and restricts the performance of continual pre-training.
Towards this issue, we investigate domain-adaptive language understanding with a limited target corpus, and propose a novel language modeling method named VarMAE.
The method performs a context uncertainty learning module to produce diverse and well-formed contextual representations,
and can be efficiently
adapted to new domains with limited resources.

\section{Derivation of Objective Function} \label{app:loss}
Here, we take the objective for masked tokens as the example to give derivations of the loss function.
The objective for unmasked tokens is similar. For simplifying description, we omit  the superscripts that use to distinguish masked tokens from unmasked tokens.
To learn a smooth space of masked tokens where latent representations of similar contexts are close to each other and vice versa, the objective function is:
\begin{equation}
\begin{split}
    \max_{\phi, \theta} \mathbb{E}_{
    {x} \sim \mathbf{D}
    }[\mathbb{E}_{
    \mathbf{z} \sim q_{\phi}(\mathbf{z} | \mathbf{c})
    } 
    [\log p_\theta (\mathbf{x} | \mathbf{z})]],
    \\ 
    \text{s.t.} \ D_{KL}( 
            q_{\phi}(\mathbf{z} | \mathbf{c}) 
            \|
            p_{\theta}(\mathbf{z}))  < \delta,
\end{split}
\end{equation}
where $\delta>0$ is a constraint, and
$q_\phi(\mathbf{z} | \mathbf{c}) $ is the variational approximate posterior of the true posterior $p_{\theta}(\mathbf{z} | {x})$ (see Section~\ref{sec:cul}).
$D_{KL}(\cdot)$ denotes the KL-divergence term, which serves as the regularization that forces the prior distribution $p_{\theta}$ to approach the approximated posterior $q_{\phi}$.

In order to encourage this disentangling property in the inferred \cite{DBLP:conf/iclr/HigginsMPBGBML17},
we introduce a constraint $\delta$ over $q_{\phi}(\mathbf{z} | \mathbf{c})$ by matching it to a prior $p_{\theta}(\mathbf{z})$.
The objective can be computed as a Lagrangian under the KKT condition \cite{bertsekas1997nonlinear,karush2014minima}.
The above optimization problem with only one inequality constraint is equivalent to maximizing the following equation,
\begin{equation}
      \resizebox{0.86\linewidth}{!}{$
\begin{split}
    \mathcal{F}(\theta, \phi, \lambda; \mathbf{c}, \mathbf{z}) = \mathbb{E}_{
    \mathbf{z} \sim q_{\phi}(\mathbf{z} | \mathbf{c})
    }[\log p_\theta (\mathbf{x} | \mathbf{z})] 
     \\ - \lambda (D_{KL}( 
            q_{\phi}(\mathbf{z} | \mathbf{c}) 
            \|
            p_{\theta}(\mathbf{z}) ) - \delta),
\end{split}
$}
\end{equation}
where the KKT multiplier $\lambda$ is the regularization coefficient that constrains the capacity of the latent information channel $\mathbf{z}$ and puts implicit independence pressure on the learnt posterior due to the isotropic nature of the Gaussian prior $p_{\theta}(\mathbf{z})$.
Since $\delta,\lambda>0$, the function is further defined as,
\begin{equation}
    \resizebox{0.86\linewidth}{!}{$
    \begin{split}
      \mathcal{F}(\theta, \phi, \lambda; \mathbf{c}, \mathbf{z})  
      &\geq \mathcal{L}(\theta, \phi; \mathbf{c}, \mathbf{z}, \lambda)  \\ 
      &= 
      \mathbb{E}_{
      \mathbf{z} \sim q_{\phi}(\mathbf{z} | \mathbf{c})
      }[\log p_\theta (\mathbf{x} | \mathbf{z})] 
    \\ & \ \ - \lambda  D_{KL}( 
            q_{\phi}(\mathbf{z} | \mathbf{c}) 
            \|
           p_{\theta}(\mathbf{z})
           ),
    \end{split}
    $}
\end{equation}
where the multiplier $\lambda$ can be considered as a hyperparameter.
$\lambda$ not only encourages more efficient latent encoding but also creates a trade-off between context reconstruction quality and the extent of disentanglement.
We train the model by minimizing the loss $\mathcal{L}$ to push up its evidence lower bound.

\begin{table*}[t]
    \centering
      \resizebox{\linewidth}{!}{$
    \begin{tabular}{l|l|c|c|c|c|c|c|c|c|c|c} 
    \hline
    \multicolumn{1}{c|}{\multirow{3}{*}{\textbf{Corpus Size}}} 
    & \multicolumn{1}{c|}{\multirow{3}{*}{\textbf{Model}}} 
    & \multicolumn{5}{c|}{{\textit{Science-domain}}}
    & \multicolumn{5}{c}{{\textit{Finance-domain}}}    \\ \cline{3-12}
    & & \textit{ACL-ARC}  & \textit{SciCite}    & \textit{JNLPBA}   & \textit{EBM-NLP}  &  \multirow{2}{*}{{Avg.}}
    & \textit{OIR}      & \textit{MTC}      & \textit{IEE}           & \textit{PSM}      & \multirow{2}{*}{{Avg.}} 
    \\ \cline{3-6} \cline{8-11}
    & &  \multicolumn{2}{c|}{CLS}     
    &   NER      
    &   SE 
    & 
    &  \multicolumn{2}{c|}{CLS}              
    &   NER      
    &   TM              
    &   \\    \hline 
$|\mathcal{D}|/3$  & DAPT & 72.42   &   	85.92   &   	73.38   &   	75.35   &   	76.77   &   	72.65   &   	47.09   &   	66.13   &   	52.38   &   	59.56 \\ 
$|\mathcal{D}|/3$  & VarMAE & 76.98   &   	84.67   &   	74.73   &   	74.91   &   	77.82   &     	70.50   &   	53.93   &   	67.72   &   	56.02   &   	62.04 \\ \hline
$|\mathcal{D}|$  &  DAPT
&      70.02	&		84.20	&		73.85	&		75.88	&		75.99	
&	   65.54	&		54.49	&		65.90	&		46.47	&		58.10 	\\	
$|\mathcal{D}|$ & VarMAE
&      76.50	&		86.32	
&	   74.43	&		76.01
&	   \textbf{78.32}	
&	   68.77	&		56.58	
&	   70.15	&		53.68	
&	   \textbf{62.30}  
\\     \hline
    \end{tabular}
    $}
    \caption{Results of DAPT and VarMAE on all
downstream tasks against different corpus sizes of pre-training. $|\mathcal{D}|$ is the corpus size. 
    For each dataset, we run three random seeds and report the average result of the test sets.
    We report the micro-average F1 score for CLS and TM, entity-level F1 score for NER, and token-level F1 score for SE.
    }
    \label{tab:pretrain_corpus_full}
\end{table*}

\begin{table*}[t]
    \centering
      \resizebox{\linewidth}{!}{$
    \begin{tabular}{l|l|c|c|c|c|c|c|c|c|c|c} 
    \hline
    \multicolumn{1}{c|}{\multirow{3}{*}{\textbf{Masking Ratio}}} 
    & \multicolumn{1}{c|}{\multirow{3}{*}{\textbf{Model}}} 
    & \multicolumn{5}{c|}{{\textit{Science-domain}}}
    & \multicolumn{5}{c}{{\textit{Finance-domain}}}    \\ \cline{3-12}
    & & \textit{ACL-ARC}  & \textit{SciCite}    & \textit{JNLPBA}   & \textit{EBM-NLP}  &  \multirow{2}{*}{{Avg.}}
    & \textit{OIR}      & \textit{MTC}      & \textit{IEE}           & \textit{PSM}      & \multirow{2}{*}{{Avg.}} 
    \\ \cline{3-6} \cline{8-11}
    & &  \multicolumn{2}{c|}{CLS}     
    &   NER      
    &   SE 
    & 
    &  \multicolumn{2}{c|}{CLS}              
    &   NER      
    &   TM              
    &   \\    \hline 
5\% & VarMAE&  76.02   &   	85.12   &   	73.86   &   	74.09   &   	77.27   &   	67.80   &   	46.33   &   	66.72   &   	53.32   &   	58.54   \\ 
15\% & VarMAE
&      76.50	&		86.32	
&	   74.43	&		76.01
&	   \textbf{78.32}	
&	   68.77	&		56.58
&	   70.15	&		53.68	
&	   \textbf{62.30} 
\\   
30\% & VarMAE& 73.62   &   	85.69   &   	73.75   &   	74.73   &   	76.95   &   	70.57   &   	45.68   &   	65.00   &   	55.23   &   	59.12  \\ 
\hline
    \end{tabular}
    $}
    \caption{Results of VarMAE on all downstream tasks against different masking ratios of pre-training.
    For each dataset, we run three random seeds and report the average result of the test sets.
    We report the micro-average F1 score for CLS and TM, entity-level F1 score for NER, and token-level F1 score for SE.
    }
    \label{tab:pretrain_mask_full}
\end{table*}

\section{Detailed Experimental Setup} \label{sec:appendix_exper_setups}
\subsection{Datasets of Downstream Tasks}
\label{sec:appendix_dataset}
The statistics of datasets and their corresponding tasks are reported in Table~\ref{tab:datasets_sci_fi}.
\paragraph{Science Domain}
We choose four public benchmark datasets from the science domain.

\textbf{ACL-ARC} \cite{DBLP:journals/tacl/JurgensKHMJ18}
is a dataset of citation intents based on a
sample of papers from the ACL Anthology Reference Corpus \cite{DBLP:conf/lrec/BirdDDGJKLPRT08} in the NLP field. 

\textbf{SciCite} \cite{DBLP:conf/naacl/CohanAZC19} is a dataset of citation intents. 
It provides coarse-grained categories and covers a variety of scientific domains.

\textbf{JNLPBA} \cite{DBLP:conf/bionlp/CollierK04}
is a named entity dataset in the biomedical field and is derived from five superclasses in the GENIA corpus \cite{kim2003genia}.

\textbf{EBM-NLP} \cite{DBLP:conf/acl/NenkovaLYMWNP18} annotates PICO (Participants, Interventions, Comparisons and Outcomes) spans in clinical trial abstracts. The corresponding PICO Extraction task aims to identify the spans in clinical trial abstracts that describe the respective PICO elements.

\paragraph{Finance Domain}
We choose four real-world business datasets\textsuperscript{\ref{code-life}} from the 
financial domain.

\textbf{OIR} is a dataset of the outbound intent recognition task. It aims to identify the intent of customer response in the outbound call scenario. 

\textbf{MTC} is a dataset of the multi-label topic classification task. It aims to identify the topics of the spoken dialogue. 

\textbf{PSM} is a dataset of the pairwise search matching task. It aims to identify the semantic similarity of a sentence pair in the search scenario.

\textbf{IEE} is a dataset of the Insurance-entity extraction task. Its goal is to locate named entities mentioned in the input sentence.

For OIR and MTC, we use an ASR (automatic speech recognition) tool to convert acoustic signals into textual sequences in the pre-processing phase.

\subsection{Implementation Details} \label{sec:appendix_exp}

\subsubsection{Pre-training Hyperparameters} \label{sec:appendix_exp_pretrain}
Table~\ref{tab:app_params} describes the hyperparameters for pre-training on a domain-specific corpus.

\subsubsection{Fine-tuning Hyperparameters}  \label{sec:appendix_exp_fine}
Table~\ref{tab:app_params_fine} reports the fine-tuning hyperparameters for downstream tasks.

\section{Detailed Results} \label{sec:appendix_result}

In this part, we provide detailed results on science- and finance-domain downstream tasks.

\subsection{Results Against Different Corpus Sizes} \label{sec:appendix_corpus}
The detailed results of DAPT and VarMAE on all downstream tasks against different corpus sizes of pre-training are reported in Table~\ref{tab:pretrain_corpus_full}.

\subsection{Results Against Different Masking Ratios} \label{sec:appendix_mask}
{The detailed results of VarMAE on all downstream tasks against different masking ratios of pre-training are reported in Table~\ref{tab:pretrain_mask_full}.}

\section{Contribution and Future Work}
The main contributions of this work are as follows:
\textbf{1)} 
We present a domain-adaptive language modeling method named VarMAE based on the combination of variational autoencoders and masked autoencoders.
\textbf{2)} 
We design a context uncertainty learning module to model the point-estimate context of each token into a smooth latent distribution. 
The module can produce diverse and well-formed contextual representations. 
\textbf{3)}
Extensive experiments on science- and finance-domain NLU tasks demonstrate that VarMAE can be efficiently adapted to new domains with limited resources.

For future works, we will build domain-specific structured knowledge to further assist language understanding, and apply our method for domain-adaptive language generation.

\end{document}